\documentclass{article}

\usepackage{genai4health_2026}
\usepackage{amsmath}
\usepackage{graphicx}
\usepackage{tabularx}
\usepackage{array}
\usepackage{pifont}
\usepackage{subcaption}
\usepackage{float}
\usepackage[utf8]{inputenc} 
\usepackage[T1]{fontenc}    
\usepackage{hyperref}       
\usepackage{url}            
\usepackage{booktabs}       
\usepackage{amsfonts}       
\usepackage{nicefrac}       
\usepackage{microtype}      
\usepackage{xcolor}         

\newcommand{\cmark}{\ding{51}}
\newcommand{\xmark}{\ding{55}}

\title{Synthetic EHR Generators Should Be Audited Like LLMs}

\author{%
  Inigo Stibbe\textsuperscript{1} \quad
  Padmanabhan Ramnarayan\textsuperscript{2} \quad
  A. Aldo Faisal\textsuperscript{1,3} \\[0.5em]
  \textsuperscript{1}Departments of Computing and Bioengineering,
  Imperial College London, UK \\
  \textsuperscript{2}Department of Surgery \& Cancer,
  Imperial College London, UK \\
  \textsuperscript{3}Chair in Digital Health \& Data Science,
  University of Bayreuth, Germany \\[0.5em]
  {\small\texttt{\{i.stibbe25,p.ramnarayan,a.faisal\}@imperial.ac.uk}}
}

\begin{document}

\maketitle

\vspace{-4mm}

\begin{abstract} 

Synthetic electronic health record (EHR) generation is increasingly adopting large autoregressive models, yet privacy evaluation has not kept pace, relying heavily on similarity-based heuristics and aggregate metrics that can obscure patient-specific memorisation. We argue that these models should additionally undergo LLM-style memorisation audits at the level of individual patient records. We support this position by comparing membership-inference evaluations across four prominent autoregressive EHR generators, highlighting inconsistent evaluation practices in patient representations and attack procedures. We evaluate trajectory reconstruction from partial context and introduce randomly assigned patient-level canaries to distinguish memorised associations from clinical generalisation. We examine these methods in a GPT-2-style model deliberately overfit to 1,000 MIMIC-IV patient trajectories, creating a controlled setting in which memorisation should be readily detectable. The model systematically recovers training patients' canaries, and their trajectory reconstruction approaches perfect accuracy with increasing context. Held-out patients show no systematic canary recovery, while reconstruction remains low and improves only marginally with additional context. These experiments establish a positive-control baseline for further evaluation of these methods in realistically trained models.

\end{abstract}

\vspace{-3mm}

\section{Introduction}

Electronic health records (EHRs) support a wide range of clinical and research applications, but access remains restricted by concerns about patient privacy, institutional governance, and regulatory compliance. Synthetic data has therefore emerged as a potential means of enabling wider data use without
directly sharing identifiable patient records.

However, establishing that synthetic EHRs adequately protect patient privacy remains difficult. Regulations such as the GDPR and HIPAA establish requirements for anonymisation and de-identification, but do not prescribe a standardised technical framework for evaluating synthetic data produced by modern generative models. Recent synthetic-data guidance has similarly highlighted the need for clearly defined disclosure metrics and decision thresholds \cite{pilgram_2025_consensus}.

Recent synthetic EHR generators increasingly model longitudinal patient records autoregressively, factorising each record into a series of conditional predictions. This is the same modelling principle that underpins LLMs, in which each element is predicted from the preceding context. While well suited to ordered, variable-length patient histories, it introduces an analogous memorisation risk: the model may reproduce a training patient's record when provided with its preceding clinical context.
    
\textbf{Central position: We argue that synthetic EHR generators should undergo LLM-style memorisation audits that quantify leakage at the level of individual patient records.}

We first compare the privacy evaluations used by four prominent autoregressive EHR generators, showing that nominally similar assessments differ substantially in their representations and attack procedures. We then introduce a canary-based proxy for patient-level memorisation and evaluate its behaviour on an intentionally overfit autoregressive model.

\section{Background and Related Work}

\subsection{Synthetic EHR Generation}

Early approaches to generate synthetic EHRs relied on rule-based and statistical methods \cite{buczak_2010_datadriven, walonoski_2018_synthea}. Deep generative modelling was later introduced by methods such as medGAN \cite{choi_2017_generating}, followed by architectures including variational autoencoders and diffusion models \cite{biswal_2021_eva, naseer_2023_scoehr, yuan_2024_ehrdiff}.

EHRs contain heterogeneous clinical events distributed across irregularly spaced visits and multiple tables. Representing each patient as an ordered sequence of clinical-event tokens has therefore become increasingly common \cite{wang_2022_promptehr, theodorou_2023_synthesize, pang_2024_cehrgpt, cho_2025_generating}. Autoregressive models can operate directly on these sequences, preserving temporal order, repeated events, and variable-length patient histories.

\subsection{Privacy Evaluation of Synthetic EHRs}

Privacy evaluations of synthetic data commonly examine membership disclosure, attribute disclosure, identity disclosure, and similarity between real and synthetic records. Membership inference asks whether an attacker can determine if a patient contributed to training \cite{zhang_2022_membership}; attribute inference asks whether unknown sensitive information can be inferred from known attributes; and identity disclosure considers whether a synthetic record can be linked to an identifiable individual \cite{pilgram_2025_consensus}. Nearest-neighbour adversarial accuracy (NNAA) contrasts training--synthetic and held-out--synthetic nearest-neighbour relationships to diagnose overfitting. However, recent consensus guidance cautions that record-level similarity alone is not a valid proxy for identity disclosure \cite{pilgram_2025_consensus,yao_2025_dcr}.

To examine how these evaluations are applied to autoregressive synthetic EHR generation, we review four prominent models: PromptEHR, HALO, CEHR-GPT, and RawMed \cite{wang_2022_promptehr,theodorou_2023_synthesize, pang_2024_cehrgpt,cho_2025_generating}. These models generate longitudinal patient records using representations ranging from visit-level multi-hot vectors to tokenised clinical-event sequences. We focus on them because all four report membership-inference results, enabling comparison of assessments that share the same stated objective.

\begin{table}[h]
  \caption{Privacy tests reported by prominent autoregressive synthetic-EHR
  generators. MIA: membership inference attack; AIA: attribute inference
  attack; MIDR: meaningful identity disclosure risk; NNAA:
  nearest-neighbour adversarial accuracy.}
  \label{tab:privacy-tests}
  \centering
  \small
  \setlength{\tabcolsep}{4pt}
  \renewcommand{\arraystretch}{1.15}
  \begin{tabular}{llcccc}
    \toprule
    \textbf{Model} & \textbf{Dataset} & \textbf{MIA} & \textbf{AIA} &
    \textbf{MIDR} & \textbf{NNAA} \\
    \midrule
    PromptEHR \cite{wang_2022_promptehr}
      & MIMIC-III & \cmark & \cmark & \xmark & \xmark \\
    HALO \cite{theodorou_2023_synthesize}
      & MIMIC-III + private & \cmark & \cmark & \xmark & \cmark \\
    CEHR-GPT \cite{pang_2024_cehrgpt}
      & Private & \cmark & \cmark & \cmark & \cmark \\
    RawMed \cite{cho_2025_generating}
      & MIMIC-IV + eICU & \cmark & \xmark & \xmark & \xmark \\
    \bottomrule
  \end{tabular}
\end{table}

Table~\ref{tab:privacy-tests} shows that membership inference is the only assessment reported by all four models. However, the implementations differ substantially. PromptEHR uses the predictions of a learned shadow model to train a membership classifier and reports AUROC. HALO evaluates both model likelihood and Hamming distance to generated records, classifying the closest half as members and reporting accuracy. CEHR-GPT discards event ordering by flattening each patient into a multi-hot vector, then applies a fixed threshold to Euclidean distance. RawMed instead calculates position-wise Hamming distance between fixed-size token matrices and applies a threshold to the nearest synthetic record. Further implementation details are provided in Appendix \ref{app:mia-comparison}.

Thus, assessments bearing the same ``membership inference'' label differ in their patient representations, attack signals, decision rules, and reported metrics, making results difficult to compare across models. Moreover, their conclusions are summarised using global accuracy or AUROC. Such averages may conceal substantial heterogeneity: an attack can perform poorly overall while confidently identifying a small number of rare or distinctive patients \cite{knolle_2026_disparate}. Near-random aggregate performance therefore does not establish that no patient has been memorised; it shows only that the evaluated attack performs poorly on average.

\subsection{Memorisation and Auditing in LLMs}

Memorisation is defined by Carlini et al.~\cite{carlini_2023_quantifying} as:

\begin{quote}
A string $s$ is extractable with $k$ tokens of context from a model $f$ if there exists a length-$k$ string $p$ such that the concatenation $[p \,\|\, s]$ is contained in the training data for $f$, and $f$ produces $s$ when prompted with $p$ using greedy decoding.
\end{quote}

This definition transfers naturally to autoregressive EHR models. When a patient record is represented as a sequence of clinical-event tokens, $p$ corresponds to an observed prefix of the patient's trajectory and $s$ to its continuation or some sensitive attribute. Extractable memorisation therefore occurs when the model reproduces a continuation from a training patient's record after being given the preceding clinical context.

LLM memorisation is commonly evaluated through extraction attacks, likelihood- or perplexity-based scores, and canary audits \cite{carlini_2019_secret, carlini_2021_extracting, carlini_2023_quantifying}. Extraction attacks have demonstrated that language models can reproduce verbatim training sequences, including personally identifiable information \cite{carlini_2021_extracting}. Canary audits instead insert random, training-specific sequences and measure their subsequent exposure \cite{carlini_2019_secret}. Because a random canary cannot be inferred from the underlying data distribution, its recovery provides a positive control for memorisation rather than generalisation.

\section{Synthetic EHR Generators Should Be Audited Like LLMs}

We argue that privacy evaluation for synthetic EHR generators should expand beyond aggregate dataset-level assessments to include internal, patient-level audits of the trained model. Membership-inference and nearest-neighbour metrics characterise important forms of disclosure, but their aggregate scores may conceal memorisation concentrated among a small number of patients. LLM-style auditing provides complementary tools for identifying this tail risk before synthetic data are released.

Tonekaboni et al. \cite{tonekaboni_2025_investigation} provide the closest related work, introducing six black-box tests for generative and embedding memorisation in structured EHR foundation models, including trajectory reconstruction, sensitive-attribute prediction, membership inference, and patient-level risk assessment. We build on their central premise that EHR models require patient-level memorisation audits, and argue that these audits should become a standard component of privacy evaluation for synthetic EHR generators. Our preliminary experiment explores a complementary positive control: randomly assigned patient-level canaries learned during model training, rather than memorisation introduced through a deterministic rule in a toy model.

\section{A Canary-Based Memorisation Audit for Synthetic EHR Generators}

Reconstruction of clinical-event sequences provides the most direct test of extractable memorisation, but correct reconstruction may arise from predictable clinical regularities rather than patient-specific retention. To address this confound, we introduce randomly assigned patient-level canaries as an internal audit conducted by the model developer or data custodian.

Within each split, we randomly permute a balanced collection of binary labels $l_i \in \{A,B\}$ across patient trajectories $u_i$, appending each assigned label to the corresponding trajectory, such that
\[
P(l_i=A)=P(l_i=B)=\tfrac{1}{2}.
\]
The labels are therefore independent of patient content, although assignments are not mutually independent because their totals are fixed. Systematic recovery of these arbitrary associations provides evidence of patient-specific memorisation, but does not itself demonstrate extraction of clinical events.

Let $z_{i,A}$ and $z_{i,B}$ denote the model logits for the two canary tokens after conditioning on $u_i$. We define the pairwise probability assigned to the true canary as
\[
\pi_i =
\begin{cases}
\sigma(z_{i,A}-z_{i,B}), & l_i=A,\\
\sigma(z_{i,B}-z_{i,A}), & l_i=B,
\end{cases}
\]
where $\sigma$ is the logistic sigmoid. Under the random-assignment null,
$\mathbb{E}_{l_i}[\pi_i \mid u_i]=\tfrac{1}{2}$, although individual values may differ from $\tfrac{1}{2}$.

We define the patient-level memorisation score
\begin{equation}
    \mu_i=2\pi_i-1,
\end{equation}
where $\mu_i=0$ indicates no preference between the canaries, positive values favour the assigned canary, and negative values favour the alternative. Under the null, the distribution of $\mu_i$ is expected to be approximately symmetric about zero. We therefore define an excess-tail statistic $\tau(\gamma)$ measuring whether strongly positive scores occur more frequently than equally strong negative scores (Appendix~\ref{app:metric}).

Following the ETHOS framework, we convert MIMIC-IV into the MEDS format and represent each patient record as a tokenised sequence \cite{renc_2024_zero, johnson_2023_mimiciv, mcdermott_2025_meds}. We train a 49.5M-parameter GPT-2-style model (Appendix \ref{app:model}), intended for a complete MIMIC-IV training cohort of approximately 230,000 patients after an 80/10/10 split. For this preliminary proof-of-concept, however, we deliberately restrict training to 1,000 patients and continue into the extreme overfit regime. This provides a positive-control setting in which patient-level memorisation is expected, allowing us to test whether the proposed score detects memorisation and whether the same setting also produces reconstruction from partial context. The experiment is not intended to estimate memorisation under a realistic full-cohort training regime.

\begin{figure}[h]
  \centering
  \begin{subfigure}[b]{0.48\textwidth}
    \centering
    \includegraphics[width=\textwidth]{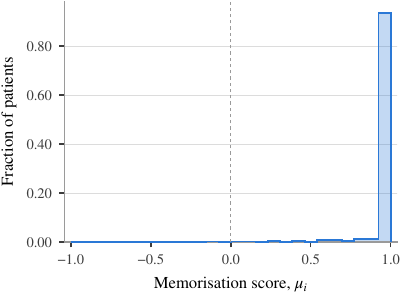}
    \caption{Training patients}
    \label{fig:canary_trained}
  \end{subfigure}
  \hfill
  \begin{subfigure}[b]{0.48\textwidth}
    \centering
    \includegraphics[width=\textwidth]{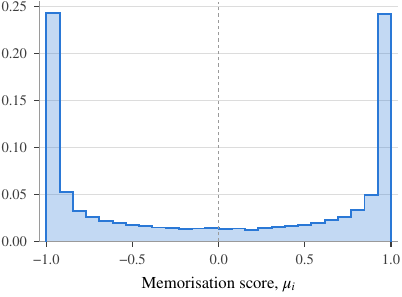}
    \caption{Held-out patients}
    \label{fig:canary_heldout}
  \end{subfigure}
  
  \caption{Patient-level memorisation scores for the deliberately overfit model. Training: \(n=1{,}000\), \(\bar{\mu}=0.969\), \(\tau_{0.95}=+0.925\). Held-out: \(n=28{,}560\), \(\bar{\mu}=-0.002\), \(\tau_{0.95}=-0.001\).}
  \label{fig:canary_arms_comparison}
\end{figure}

Figure~\ref{fig:canary_arms_comparison} shows the distribution of memorisation scores for the 1,000 training patients and the full held-out validation set. In the deliberately overfit training cohort, scores are concentrated near $1$, indicating that the model assigns high probability to the patient-specific canary observed during training. Validation scores are instead distributed approximately symmetrically between positive and negative values: the model often expresses high confidence, but is equally likely to favour the correct or incorrect randomly assigned canary. Consequently, the mean validation score remains close to zero, as expected when the random canary assignment has not been memorised.

\begin{figure}[h]
  \centering
  \begin{subfigure}[b]{0.48\textwidth}
    \centering
    \includegraphics[width=\textwidth]{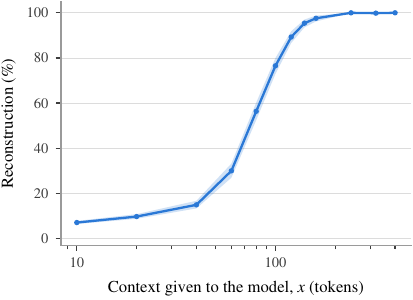}
    \caption{Training patients}
    \label{fig:recon_train}
  \end{subfigure}
  \hfill
  \begin{subfigure}[b]{0.48\textwidth}
    \centering
    \includegraphics[width=\textwidth]{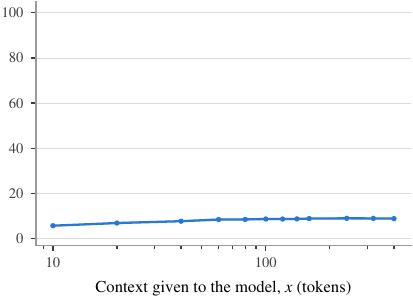}
    \caption{Held-out patients}
    \label{fig:recon_val}
    
  \end{subfigure}
  
  \caption{Fixed-target reconstruction with greedy decoding: mean proportion of tokens 401--500 recovered given the immediately preceding \(x\) ground-truth tokens. Trajectories of \(\geq 500\) tokens only (\(n=406\) train, \(1{,}000\) validation). Bands show 95\% patient-level bootstrap CIs.}
  \label{fig:recon}
\end{figure}

Systematic recovery of the assigned canaries demonstrates that the model has retained an arbitrary patient-specific association, but does not by itself establish that clinical events from the corresponding training records are extractable. We therefore evaluate trajectory reconstruction in Figure \ref{fig:recon}. Reconstruction improves with additional context for training patients and eventually approaches perfect accuracy, whereas performance remains substantially lower for held-out patients and improves only slightly. Together, these experiments establish a controlled positive baseline: the proposed canary audit detects patient-specific associations in a setting where the model also directly reproduces training trajectories. Because canary scores saturate in this deliberately overfit regime, this
setting is not informative about their patient-level association with
reconstruction accuracy. Having established the expected positive-control
behaviour, this relationship can next be evaluated under realistic training
regimes in which the scores retain sufficient variation.

\vspace{-10pt}

\section{Alternative Views and Rebuttals}

\textbf{Will LLM-style auditing materially affect data-release decisions?} Memorisation auditing cannot by itself establish that a synthetic dataset is legally anonymous or safe to release; such decisions also depend on regulatory, institutional, and governance considerations. However, aggregate similarity or membership-inference scores may conceal risks concentrated among individual patients. Patient-level auditing can provide data custodians with more transparent evidence about this tail risk before release. Given that the lack of standardised privacy evaluation is itself a barrier to the adoption of synthetic medical data \cite{kaabachi_2025_scoping}, we view LLM-style auditing not as a guarantee of approval, but as a means of providing a clearer and more consistent evidential basis for release decisions. 

\textbf{Why use artificial canaries when trajectory reconstruction measures memorisation directly?} Trajectory reconstruction provides the most direct evidence that clinical information from a training record is extractable. However, correct reconstruction may arise from predictable clinical regularities rather than patient-specific memorisation, and its outcome depends on the chosen context, prediction horizon, and decoding procedure. Random canaries address this ambiguity by providing a signal that is independent of the clinical trajectory. Systematic recovery of the assigned canary therefore demonstrates retention of a patient-specific training association that cannot be explained by clinical predictability. The canary is not intended to replace reconstruction, but to provide a controlled and scalable proxy whose relationship with clinically meaningful extraction can be evaluated.

\section{Conclusion}

Synthetic EHR generation is shifting towards large autoregressive models, and privacy evaluation should adapt accordingly. We argue that these generators should undergo LLM-style memorisation audits at the level of individual patient records. As a preliminary step, we introduce a per-patient proxy based on the prediction of a canary label assigned independently of the clinical content, designed to distinguish retained patient-specific associations from predictions arising from clinical regularities. In a deliberately overfit positive-control setting, the audit shows systematic recovery of the assigned canaries for training patients, while held-out predictions follow the random-label null. Separately, trajectory-reconstruction experiments show that training records are substantially more extractable than held-out records.

These experiments establish the expected behaviour of both signals in a controlled setting, but do not yet establish their patient-level relationship. This baseline provides a foundation for developing appropriate null distributions and decision thresholds, evaluating the audit in realistically trained models, and determining how canary scores relate to clinically meaningful reconstruction and privacy risk.

\begin{ack}
Inigo Stibbe is supported by the UKRI AI Centre for Doctoral Training in Digital Healthcare (EP/Y030974/1). Padmanabhan Ramnarayan is supported by the National Institute for Health and Care Research (NIHR Research Professorship, NIHR306356). A.~Aldo Faisal acknowledges support from the UKRI AI programme and the Engineering and Physical Sciences Research Council (EPSRC) for the AI Hub in Generative Models (EP/Y028805/1).
\end{ack}


\newpage

\bibliographystyle{plain}
\bibliography{references}

@inproceedings{biswal_2021_eva,
  title = {{{EVA}}: {{Generating Longitudinal Electronic Health Records Using Conditional Variational Autoencoders}}},
  shorttitle = {{{EVA}}},
  booktitle = {Proceedings of the 6th {{Machine Learning}} for {{Healthcare Conference}}},
  author = {Biswal, Siddharth and Ghosh, Soumya and Duke, Jon and Malin, Bradley and Stewart, Walter and Xiao, Cao and Sun, Jimeng},
  year = 2021,
  month = oct,
  pages = {260--282},
  publisher = {PMLR},
  issn = {2640-3498},
  urldate = {2026-09-05},
  langid = {english}
}

@article{buczak_2010_datadriven,
  title = {Data-Driven Approach for Creating Synthetic Electronic Medical Records},
  author = {Buczak, Anna L. and Babin, Steven and Moniz, Linda},
  year = 2010,
  month = oct,
  journal = {BMC Medical Informatics and Decision Making},
  volume = {10},
  pages = {59},
  issn = {1472-6947},
  doi = {10.1186/1472-6947-10-59},
  urldate = {2026-09-03},
  langid = {english}
}

@inproceedings{carlini_2019_secret,
  title = {The Secret Sharer: Evaluating and Testing Unintended Memorization in Neural Networks},
  booktitle = {Proceedings of the 28th {{USENIX}} Conference on Security Symposium},
  author = {Carlini, Nicholas and Liu, Chang and Erlingsson, {\'U}lfar and Kos, Jernej and Song, Dawn},
  year = 2019,
  series = {{{SEC}}'19},
  pages = {267--284},
  publisher = {USENIX Association},
  address = {USA},
  isbn = {978-1-939133-06-9}
}

@inproceedings{carlini_2021_extracting,
  title = {Extracting Training Data from Large Language Models},
  booktitle = {30th {{USENIX}} Security Symposium ({{USENIX}} Security 21)},
  author = {Carlini, Nicholas and Tram{\`e}r, Florian and Wallace, Eric and Jagielski, Matthew and {Herbert-Voss}, Ariel and Lee, Katherine and Roberts, Adam and Brown, Tom and Song, Dawn and Erlingsson, {\'U}lfar and Oprea, Alina and Raffel, Colin},
  year = 2021,
  month = aug,
  pages = {2633--2650},
  publisher = {USENIX Association},
  isbn = {978-1-939133-24-3}
}

@inproceedings{carlini_2023_quantifying,
  title = {Quantifying Memorization across Neural Language Models},
  booktitle = {The Eleventh International Conference on Learning Representations},
  author = {Carlini, Nicholas and Ippolito, Daphne and Jagielski, Matthew and Lee, Katherine and Tramer, Florian and Zhang, Chiyuan},
  year = 2023
}

@inproceedings{cho_2025_generating,
  title = {Generating Multi-Table Time Series {{EHR}} from Latent Space with Minimal Preprocessing},
  booktitle = {Advances in Neural Information Processing Systems},
  author = {Cho, Eunbyeol and Kim, Jiyoun and Lee, Minjae and Park, Sungjin and Choi, Edward},
  editor = {Belgrave, D. and Zhang, C. and Lin, H. and Pascanu, R. and Koniusz, P. and Ghassemi, M. and Chen, N.},
  year = 2025,
  volume = {38, Main Conference},
  pages = {26803--26838},
  publisher = {Curran Associates, Inc.},
  doi = {10.52202/085713-0903}
}

@inproceedings{choi_2017_generating,
  title = {Generating Multi-Label Discrete Patient Records Using Generative Adversarial Networks},
  booktitle = {Proceedings of the 2nd Machine Learning for Healthcare Conference},
  author = {Choi, Edward and Biswal, Siddharth and Malin, Bradley and Duke, Jon and Stewart, Walter F. and Sun, Jimeng},
  editor = {{Doshi-Velez}, Finale and Fackler, Jim and Kale, David and Ranganath, Rajesh and Wallace, Byron and Wiens, Jenna},
  year = 2017,
  month = aug,
  series = {Proceedings of Machine Learning Research},
  volume = {68},
  pages = {286--305},
  publisher = {PMLR}
}

@article{johnson_2023_mimiciv,
  title = {{{MIMIC-IV}}, a Freely Accessible Electronic Health Record Dataset},
  author = {Johnson, Alistair E. W. and Bulgarelli, Lucas and Shen, Lu and Gayles, Alvin and Shammout, Ayad and Horng, Steven and Pollard, Tom J. and Hao, Sicheng and Moody, Benjamin and Gow, Brian and Lehman, Li-wei H. and Celi, Leo A. and Mark, Roger G.},
  year = 2023,
  month = jan,
  journal = {Scientific Data},
  volume = {10},
  number = {1},
  pages = {1},
  issn = {2052-4463},
  doi = {10.1038/s41597-022-01899-x}
}

@article{kaabachi_2025_scoping,
  title = {A Scoping Review of Privacy and Utility Metrics in Medical Synthetic Data},
  author = {Kaabachi, Bayrem and Despraz, J{\'e}r{\'e}mie and Meurers, Thierry and Otte, Karen and Halilovic, Mehmed and Kulynych, Bogdan and Prasser, Fabian and Raisaro, Jean Louis},
  year = 2025,
  month = jan,
  journal = {npj Digital Medicine},
  volume = {8},
  number = {1},
  pages = {60},
  issn = {2398-6352},
  doi = {10.1038/s41746-024-01359-3}
}

@article{knolle_2026_disparate,
  title = {Disparate Privacy Risks from Medical {{AI}}},
  author = {Knolle, Moritz A. and Menten, Martin J. and Jungmann, Friederike and Meissen, Felix and Glocker, Ben and Rueckert, Daniel and Kaissis, Georgios},
  year = 2026,
  month = aug,
  journal = {Nature},
  volume = {656},
  number = {8126},
  pages = {192--198},
  issn = {1476-4687},
  doi = {10.1038/s41586-026-10688-0}
}

@inproceedings{mcdermott_2025_meds,
  title = {{{MEDS}}: {{Building}} Models and Tools in a Reproducible Health {{AI}} Ecosystem},
  booktitle = {Proceedings of the 31st {{ACM SIGKDD}} Conference on Knowledge Discovery and Data Mining {{V}}.2},
  author = {McDermott, Matthew B. A. and Xu, Justin and Bergamaschi, Teya S. and Jeong, Hyewon and Lee, Simon A. and Oufattole, Nassim and Rockenschaub, Patrick and Stankevi{\v c}i{\=u}t{\.e}, Kamil{\.e} and Steinberg, Ethan and Sun, Jimeng and van de Water, Robin P. and Wornow, Michael and Wu, John and Wu, Zhenbang},
  year = 2025,
  series = {Kdd '25},
  pages = {6243--6244},
  publisher = {Association for Computing Machinery},
  address = {New York, NY, USA},
  doi = {10.1145/3711896.3737608},
  isbn = {979-8-4007-1454-2}
}

@inproceedings{naseer_2023_scoehr,
  title = {{{ScoEHR}}: {{Generating}} Synthetic Electronic Health Records Using Continuous-Time Diffusion Models},
  booktitle = {Proceedings of the 8th Machine Learning for Healthcare Conference},
  author = {Naseer, Ahmed Ammar and Walker, Benjamin and Landon, Christopher and Ambrosy, Andrew and Fudim, Marat and Wysham, Nicholas and Toro, Botros and Swaminathan, Sumanth and Lyons, Terry},
  editor = {Deshpande, Kaivalya and Fiterau, Madalina and Joshi, Shalmali and Lipton, Zachary and Ranganath, Rajesh and Urteaga, I{\~n}igo and Yeung, Serene},
  year = 2023,
  month = aug,
  series = {Proceedings of Machine Learning Research},
  volume = {219},
  pages = {489--508},
  publisher = {PMLR}
}

@article{pang_2024_cehrgpt,
  title = {{{CEHR-GPT}}: {{Generating}} Electronic Health Records with Chronological Patient Timelines},
  author = {Pang, Chao and Jiang, Xinzhuo and Pavinkurve, Nishanth Parameshwar and Kalluri, Krishna S. and Minto, Elise L. and Patterson, Jason and Zhang, Linying and Hripcsak, George and G{\"u}rsoy, Gamze and Elhadad, No{\'e}mie and Natarajan, Karthik},
  year = 2024,
  journal = {arXiv preprint arXiv:2402.04400},
  eprint = {2402.04400},
  primaryclass = {cs.LG},
  doi = {10.48550/arXiv.2402.04400},
  archiveprefix = {arXiv}
}

@article{pilgram_2025_consensus,
  title = {A Consensus Privacy Metrics Framework for Synthetic Data},
  author = {Pilgram, Lisa and Dankar, Fida Kamal and Drechsler, J{\"o}rg and Elliot, Mark and {Domingo-Ferrer}, Josep and Francis, Paul and Kantarcioglu, Murat and Kong, Linglong and Malin, Bradley and Muralidhar, Krishnamurty and Myles, Puja and Prasser, Fabian and Raisaro, Jean Louis and Yan, Chao and El Emam, Khaled},
  year = 2025,
  journal = {Patterns},
  volume = {6},
  number = {10},
  pages = {101320},
  issn = {2666-3899},
  doi = {10.1016/j.patter.2025.101320}
}

@article{renc_2024_zero,
  title = {Zero Shot Health Trajectory Prediction Using Transformer},
  author = {Renc, Pawel and Jia, Yugang and Samir, Anthony E. and Was, Jaroslaw and Li, Quanzheng and Bates, David W. and Sitek, Arkadiusz},
  year = 2024,
  month = sep,
  journal = {npj Digital Medicine},
  volume = {7},
  number = {1},
  pages = {256},
  issn = {2398-6352},
  doi = {10.1038/s41746-024-01235-0}
}

@article{theodorou_2023_synthesize,
  title = {Synthesize High-Dimensional Longitudinal Electronic Health Records via Hierarchical Autoregressive Language Model},
  author = {Theodorou, Brandon and Xiao, Cao and Sun, Jimeng},
  year = 2023,
  month = aug,
  journal = {Nature Communications},
  volume = {14},
  number = {1},
  pages = {5305},
  issn = {2041-1723},
  doi = {10.1038/s41467-023-41093-0}
}

@inproceedings{tonekaboni_2025_investigation,
  title = {An Investigation of Memorization Risk in Healthcare Foundation Models},
  booktitle = {Advances in Neural Information Processing Systems},
  author = {Tonekaboni, Sana and Stempfle, Lena and Fallahpour, Adibvafa and Gerych, Walter and Ghassemi, Marzyeh},
  editor = {Belgrave, D. and Zhang, C. and Lin, H. and Pascanu, R. and Koniusz, P. and Ghassemi, M. and Chen, N.},
  year = 2025,
  volume = {38, Main Conference},
  pages = {10555--10580},
  publisher = {Curran Associates, Inc.},
  doi = {10.52202/085713-0353}
}

@article{walonoski_2018_synthea,
  title = {Synthea: {{An}} Approach, Method, and Software Mechanism for Generating Synthetic Patients and the Synthetic Electronic Health Care Record},
  author = {Walonoski, Jason and Kramer, Mark and Nichols, Joseph and Quina, Andre and Moesel, Chris and Hall, Dylan and Duffett, Carlton and Dube, Kudakwashe and Gallagher, Thomas and McLachlan, Scott},
  year = 2018,
  month = mar,
  journal = {Journal of the American Medical Informatics Association},
  volume = {25},
  number = {3},
  eprint = {https://academic.oup.com/jamia/article-pdf/25/3/230/34150150/ocx079.pdf},
  pages = {230--238},
  issn = {1527-974X},
  doi = {10.1093/jamia/ocx079}
}

@inproceedings{wang_2022_promptehr,
  title = {{{PromptEHR}}: {{Conditional}} Electronic Healthcare Records Generation with Prompt Learning},
  booktitle = {Proceedings of the 2022 Conference on Empirical Methods in Natural Language Processing},
  author = {Wang, Zifeng and Sun, Jimeng},
  editor = {Goldberg, Yoav and Kozareva, Zornitsa and Zhang, Yue},
  year = 2022,
  month = dec,
  pages = {2873--2885},
  publisher = {Association for Computational Linguistics},
  address = {Abu Dhabi, United Arab Emirates},
  doi = {10.18653/v1/2022.emnlp-main.185}
}

@inproceedings{yao_2025_dcr,
  title = {The {{DCR}} Delusion: {{Measuring}} the Privacy Risk of Synthetic Data},
  booktitle = {Computer Security -- {{ESORICS}} 2025},
  author = {Yao, Zexi and Kr{\v c}o, Nata{\v s}a and Ganev, Georgi and {de Montjoye}, Yves-Alexandre},
  editor = {Nicomette, Vincent and Benzekri, Abdelmalek and {Boulahia-Cuppens}, Nora and Vaidya, Jaideep},
  year = 2025,
  pages = {469--487},
  publisher = {Springer Nature Switzerland},
  address = {Cham},
  isbn = {978-3-032-07884-1}
}

@article{yuan_2024_ehrdiff,
  title = {{{EHRDiff}} : {{Exploring}} Realistic {{EHR}} Synthesis with Diffusion Models},
  author = {Yuan, Hongyi and Zhou, Songchi and Yu, Sheng},
  year = 2024,
  journal = {Transactions on Machine Learning Research},
  issn = {2835-8856}
}

@article{zhang_2022_membership,
  title = {Membership Inference Attacks against Synthetic Health Data},
  author = {Zhang, Ziqi and Yan, Chao and Malin, Bradley A.},
  year = 2022,
  journal = {Journal of Biomedical Informatics},
  volume = {125},
  pages = {103977},
  issn = {1532-0464},
  doi = {10.1016/j.jbi.2021.103977}
}


\medskip

{
\small

\newpage

\appendix

\section{Comparison of Membership-Inference Evaluations}
\label{app:mia-comparison}

\begin{figure}[h]
    \centering
    \includegraphics[width=1\linewidth]{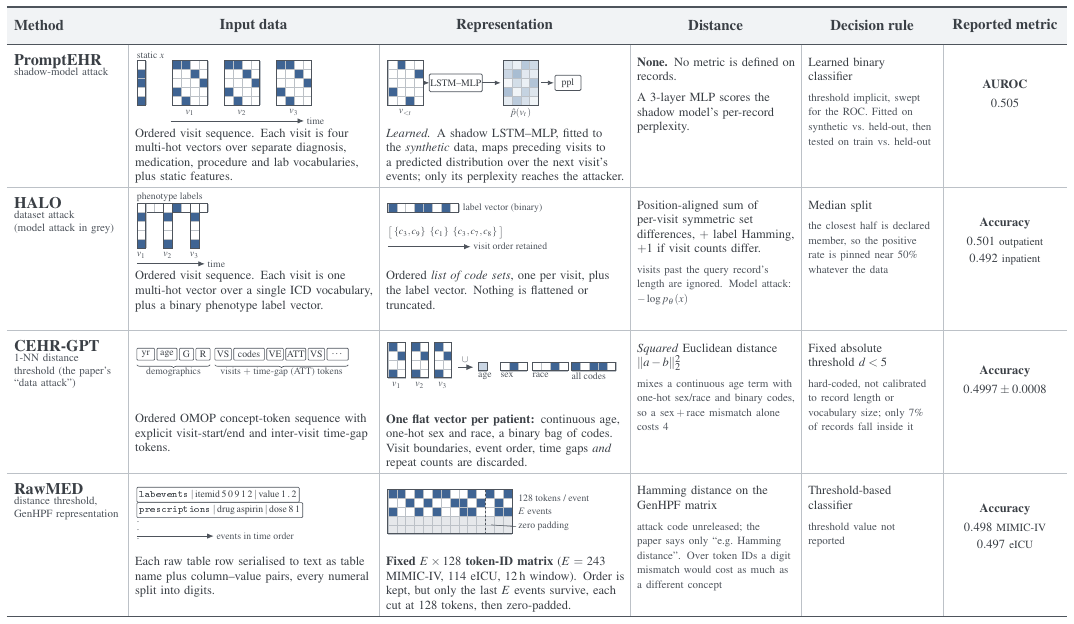}
    \caption{Comparison of membership-inference implementations across four autoregressive synthetic-EHR generators. The visualisations are schematic and illustrate the data representations used for evaluation, not the generators' exact model architectures.}
    \label{fig:placeholder}
\end{figure}

\section{Excess-Tail Memorisation Statistic}
\label{app:metric}

For a threshold $\gamma \in [0,1]$, we define the positive- and negative-tail
rates as
\begin{equation}
    T^{+}(\gamma)
    =
    \frac{1}{N}\sum_{i=1}^{N}
    \mathbb{I}\{\mu_i > \gamma\},
    \qquad
    T^{-}(\gamma)
    =
    \frac{1}{N}\sum_{i=1}^{N}
    \mathbb{I}\{\mu_i < -\gamma\}.
\end{equation}

Because the canary labels are assigned independently of the patient
trajectories, the distribution of $\mu_i$ should be approximately symmetric
about zero when the patient--canary associations have not been memorised.
We therefore define the excess-tail memorisation statistic as
\begin{equation}
    \tau(\gamma)
    =
    T^{+}(\gamma)-T^{-}(\gamma)
    =
    \frac{1}{N}\sum_{i=1}^{N}
    \left[
        \mathbb{I}\{\mu_i>\gamma\}
        -
        \mathbb{I}\{\mu_i<-\gamma\}
    \right].
    \label{eq:tau}
\end{equation}

Positive values of $\tau(\gamma)$ indicate an excess of patients for whom
the model strongly favours the assigned canary over patients for whom it
equally strongly favours the alternative canary. Under the no-memorisation
null, we expect $\tau(\gamma)$ to be close to zero.

\newpage
\section{Summary of Data and Model Architecture}
\label{app:model}

\begin{table}[h]
\centering
\small
\caption{\textbf{Sequence statistics for the two cohorts.} Members are the 1{,}000 records the model was trained on; non-members are the full validation split of the same canary dataset. }

\label{tab:sequences}
\begin{tabular}{@{}lrr@{}}
\toprule
& Members & Non-members \\
\midrule
Patients                            & 1,000 & 28,560 \\
Canary \texttt{A} : \texttt{B}        & 500 : 500 & 14,280 : 14,280 \\
Total tokens                        & 1,181,581 & 36,188,692 \\
\addlinespace
Length, mean $\pm$ sd               & 1,182 $\pm$ 2,728 & 1,267 $\pm$ 3,252 \\
\quad minimum                       & 9 & 7 \\
\quad 25th percentile               & 137 & 127 \\
\quad median                        & 344 & 340 \\
\quad 75th percentile               & 1,033 & 1,061 \\
\quad 95th percentile               & 4,805 & 5,206 \\
\quad 99th percentile               & 12,415 & 15,055 \\
\quad maximum                       & 40,358 & 96,689 \\
\addlinespace
Records $>$ 2{,}048 tokens            & 13.7\% & 14.2\% \\
Records $<$ 150 tokens              & 27.6\% & 29.4\% \\
\bottomrule
\end{tabular}
\end{table}

\begin{table}[h]
\centering
\small
\caption{\textbf{Configuration of the overfit model.} A GPT-2-style model fitted to 1{,}000 MIMIC-IV
patient records to induce memorisation.}
\label{tab:model-params}
\begin{tabular}{@{}lr@{}}
\toprule
\multicolumn{2}{@{}l}{\textit{Architecture}} \\
Layers                      & 6 \\
Attention heads             & 12 \\
Embedding dimension         & 768 \\
Context window              & 2{,}048 \\
Positional encoding         & RoPE ($10^4$) \\
Dropout                     & 0.1 \\
Vocabulary                  & 4{,}511 \\
Parameters                  & 49.5M \\
\addlinespace

\multicolumn{2}{@{}l}{\textit{Optimisation}} \\
Optimiser                   & AdamW \\
Learning rate               & $1.5\times10^{-4}$ \\
$\beta_1$, $\beta_2$        & 0.9, 0.95 \\
Weight decay                & 0.1 \\
Batch size                  & 64 \\
Training steps              & 10{,}000 \\
\quad warmup / hold / decay & 250 / 2{,}500 / 5{,}000 \\
Final LR ratio              & 0.1 \\
\bottomrule
\end{tabular}
\end{table}

\vspace{-2pt}

\begin{figure}[H]
  \centering
  \includegraphics[width=0.5\textwidth]{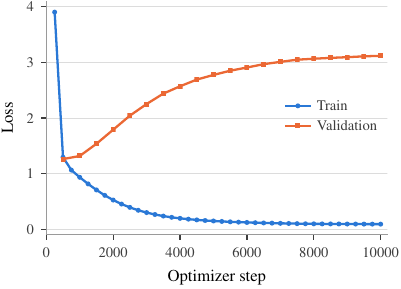}
  \caption{Training and validation loss for the deliberately overfit model (\(n=1{,}000\) training subjects, \(28{,}560\) held out).}
  \label{fig:training-curve}
\end{figure}




\newpage

} 

\end{document}